\documentclass[runningheads]{llncs}
\usepackage{graphicx}
\usepackage{amsmath,amssymb} % define this before the line numbering.
\usepackage{color}
\usepackage[width=122mm,left=12mm,paperwidth=146mm,height=193mm,top=12mm,paperheight=217mm]{geometry}
\usepackage{subfig}

\begin{document}
% \linenumbers
\pagestyle{headings}
\mainmatter

\title{Depth2Action: Exploring Embedded Depth for Large-Scale Action Recognition} % Replace with your title

\titlerunning{Depth2Action: Exploring Embedded Depth for Action Recognition}

\authorrunning{Yi Zhu and Shawn Newsam}

\author{Yi Zhu and Shawn Newsam}

%Please write out author names in full in the paper, i.e. full given and family names. 
%If any authors have names that can be parsed into FirstName LastName in multiple ways, please include the correct parsing, in a comment to the volume editors:
%\index{Lastnames, Firstnames}
%(Do not uncomment it, because you may introduce extra index items if you do that...)

\institute{University of California, Merced\\
	\email{ \{yzhu25,snewsam\}@ucmerced.edu}
}

\maketitle

\begin{abstract}
This paper performs the first investigation into depth for large-scale human action recognition in video \emph{where the depth cues are estimated from the videos themselves}. We develop a new framework called \emph{depth2action} and experiment thoroughly into how best to incorporate the depth information. We introduce spatio-temporal depth normalization (STDN) to enforce temporal consistency in our estimated depth sequences. We also propose modified depth motion maps (MDMM) to capture the subtle temporal changes in depth. These two components significantly improve the action recognition performance. We evaluate our depth2action framework on three large-scale action recognition video benchmarks. Our model achieves state-of-the-art performance when combined with appearance and motion information thus demonstrating that depth2action is indeed complementary to existing approaches.

\end{abstract}

\section{Introduction}
Human action recognition in video is a fundamental problem in computer vision due to its increasing importance for a range of applications such as analyzing human activity, video search and recommendation, complex event understanding, etc. 
Much progress has been made over the past several years by employing hand-crafted local features such as improved dense trajectories (IDT) \cite{idtfWang2013} or video representations that are learned directly from the data itself using deep convolutional neural networks (ConvNets). 
However, starting with the seminal two-stream ConvNets method \cite{twostream2014}, approaches have been limited to exploiting static visual information through frame-wise analysis and/or translational motion through optical flow or 3D ConvNets. Further increase in performance on benchmark datasets has been mostly due to the higher capacity of deeper networks~\cite{wanggoodpractice2015,tddwang2015,lessmoreMa2015,actionTransformations2015} or to recurrent neural networks which model long-term temporal dynamics \cite{beyondshort2015,hybridWu2015,BallasDeeper2015}.

\begin{figure}[t]
	\centering
	\includegraphics[width=0.9\linewidth,height=2.2in,trim=30 180 0 0,clip]{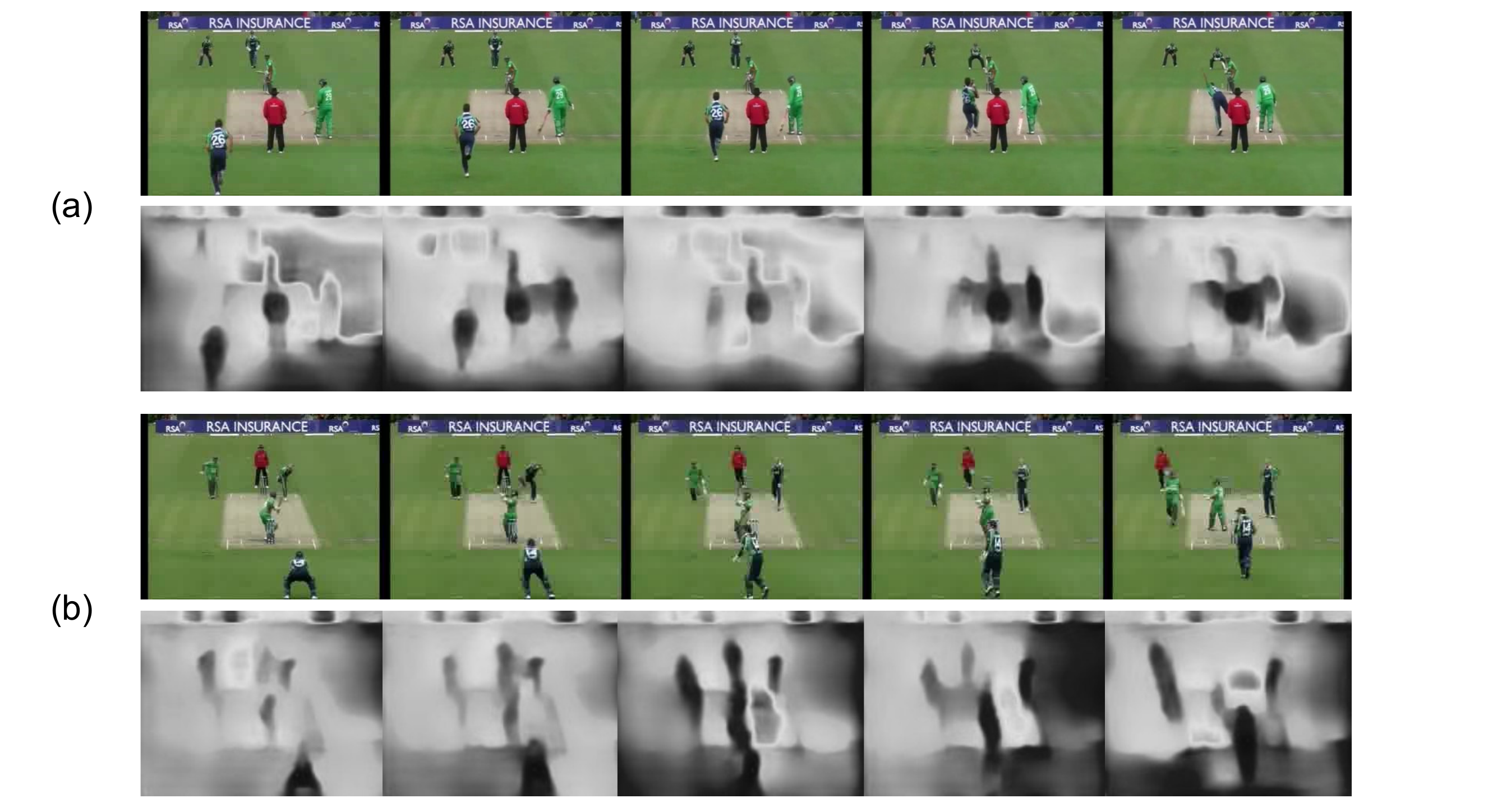}
	\caption{(a) ``CricketBowling'' and (b) ``CricketShot''. Depth information about the bowler and the batters is key to telling these two classes apart. Our proposed depth2action approach exploits the depth information that is embedded in the videos to perform large-scale action recognition. This figure is best viewed in color}
	\label{fig:depthIsUseful}
\end{figure}

Intuitively, \emph{depth} can be an important cue for recognizing complex human actions.
Depth information can help differentiate between action classes that are otherwise very similar especially with respect to appearance and translational motion in the red-green-blue (RGB) domain. 
For instance, the ``CricketShot'' and ``CricketBowling'' classes in the UCF101 dataset are often confused by the state-of-the-art models \cite{wanggoodpractice2015,actionTransformations2015}. This makes sense because, as shown in Fig. \ref{fig:depthIsUseful}, these classes can be very similar with respect to static appearance, human-object interaction, and in-plane human motion patterns. Depth information about the bowler and the batters is key to telling these two classes apart.

Previous work on depth for action recognition \cite{depthmotionmapCC2013,actiondepthWang2014,SNVdepth2014,pichaowang2015} uses depth information obtained from \emph{depth sensors} such as Kinect-like devices and thus is not applicable to large-scale action recognition in RGB video. We instead estimate the depth information \emph{directly from the video itself}. This is a difficult problem which results in noisy depth sequences and so a major contribution of our work is how to effectively extract the subtle but informative depth cues. To our knowledge, our work is the first to perform large-scale action recognition based on depth information embedded in the video data.

Our novel contributions are as follows:
(i) We introduce \emph{depth2action}, a novel approach for human action recognition using depth information embedded in videos. It is shown to be complementary to existing approaches which exploit spatial and translational motion information and, when combined with them, achieves state-of-the-art performance on three popular benchmarks.
(ii) We propose STDN to enforce temporal consistency and MDMM to capture the subtle temporal depth cues in noisy depth sequences.
(iii) We perform a thorough investigation on how best to extract and incorporate the depth cues including: image- versus video-based depth estimation; multi-stream 2D ConvNets versus 3D ConvNets to jointly extract spatial and temporal depth information; ConvNets as feature extractors versus end-to-end classifiers; early versus late fusion of features for optimal prediction; and other design choices. 

\section{Related Work}
\label{sec:related}
There exists an extensive body of literature on human action recognition. We review only the most related work.

\begin{figure}[t]
	\centering
	\includegraphics[width=1.0\linewidth,trim=0 300 0 0,clip]{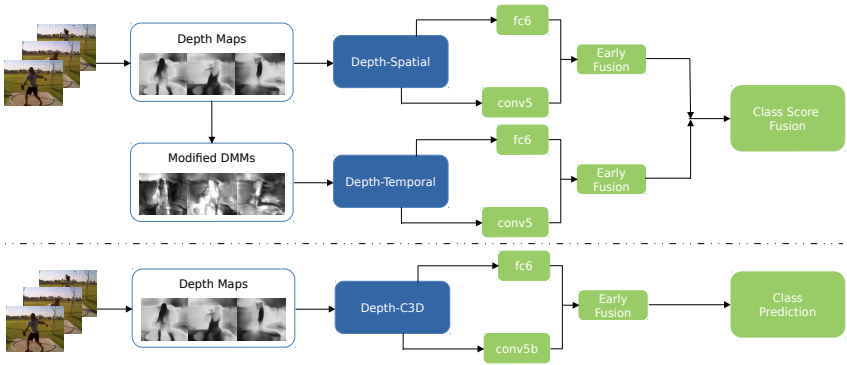}
	\caption{\textbf{Depth2Action framework}. Top: Our \emph{depth two-stream} model. Depth maps are estimated on a per-frame basis and input to a depth-spatial net. Modified depth motion maps (MDMMs) are derived from the depth maps and input to a depth-temporal net. Features are extracted, concatenated and input to two support vector machine (SVM) classifiers, to obtain the final prediction scores. Bottom: Our \emph{depth-C3D} framework which is similar except the depth maps are input to a single depth-C3D net which jointly captures spatial and temporal depth information. This figure is best viewed in color}
	\label{fig:workflow}
\end{figure}

\noindent \textbf{Deep ConvNets:} Improved dense trajectories \cite{idtfWang2013} dominated the field of video analysis for several years until the two-stream ConvNets architecture introduced by Simonyan and Zisserman \cite{twostream2014} achieved competitive results for action recognition in video. In addition, motivated by the great success of applying deep ConvNets in image analysis, researchers have adapted deep architectures to the video domain either for feature representation~\cite{tddwang2015,LCDXu2015,c3d2015,imageCNNvideo2015,unsupervisedLSTM2015} or end-to-end prediction \cite{KarpathyCVPR14,wanggoodpractice2015,beyondshort2015,hybridWu2015}.

While our framework shares some structural similarity with these works, it is distinct and complementary in that it exploits \emph{depth} for action recognition. All the works above are based on appearance and translational motion in the RGB domain. We note there has been some work that exploits audio information~\cite{learThumos2014}; however, not all videos come with audio and our approach is complementary to this work as well. 

\noindent \textbf{RGB-D Based Action Recognition:} There is previous work on action recognition in RGB-D data. Chen et al. \cite{depthmotionmapCC2013} use depth motion maps (DMM) for real-time human action recognition. Yang and Tian \cite{SNVdepth2014} cluster hypersurface normals in depth sequences to form a super normal vector (SNV) representation. Very recently, Wang et al. \cite{pichaowang2015} apply weighted hierarchical DMM and deep ConvNets to achieve state-of-the-art performance on several benchmarks.
Our work is different from approaches that use RGB-D data in several key ways:

(i) \textit{Depth information source and quality}:
These methods use depth information obtained from depth sensors. Besides limiting their applicability, this results in depth sequences that have much higher fidelity than those which can be estimated from RGB video. Our estimated depth sequences are too noisy for recognition techniques designed for depth-sensor data. Taking the difference between consecutive frames in our depth sequences only amplifies this noise making techniques such as STOP features \cite{depthSTOP2014}, SNV representations \cite{SNVdepth2014}, and DMM-based framework \cite{depthmotionmapCC2013,pichaowang2015}, for example, ineffective.

(ii) \textit{Benchmark datasets}: 
RGB-D benchmarks such as MSRAction3D~\cite{3Dactiondepth2010}, MSRDailyActivity3D~\cite{MSR3Dactivity}, MSRGesture3D~\cite{MSR3Dgesture}, MSROnlineAction3D~\cite{3DonlineAction} and MSRActionPairs3D~\cite{actionpairsHON4D2013} are much more limited in terms of the diversity of action classes and the number of samples. Further, the videos often come with other meta data like skeleton joint positions. In contrast, our benchmarks such as UCF101 contain large numbers of action classes and the videos are less constrained. Recognition is made more difficult by the large intra-class variation.

We note that we take inspiration from \cite{DMMHOG2012,pichaowang2015} in designing our modified DMMs. The approaches in these works use RGB-D data and are not appropriate for our problem, though, since they construct multiple depth sequences using different geometric projections, and our videos are too long and our estimated depth sequences too noisy to be characterized by a single DMM.

In summary, our depth2action framework is novel compared to previous work on action recognition. An overview of our framework can be found in Fig. \ref{fig:workflow}. 

\section{Methodology}
\label{sec:methodology}
Since our videos do not come with associated depth information, we need to extract it directly from the RGB video data. We consider two state-of-the-art approaches to \emph{efficiently} extract depth maps from the individual video frames. We enforce temporal consistency in these sequences through inter-frame normalization.  We explore different ConvNets architectures to extract spatial and temporal depth cues from the normalized depth sequences.

\begin{figure}[t]
	\centering
	\includegraphics[width=1.0\linewidth,trim=10 280 0 0,clip]{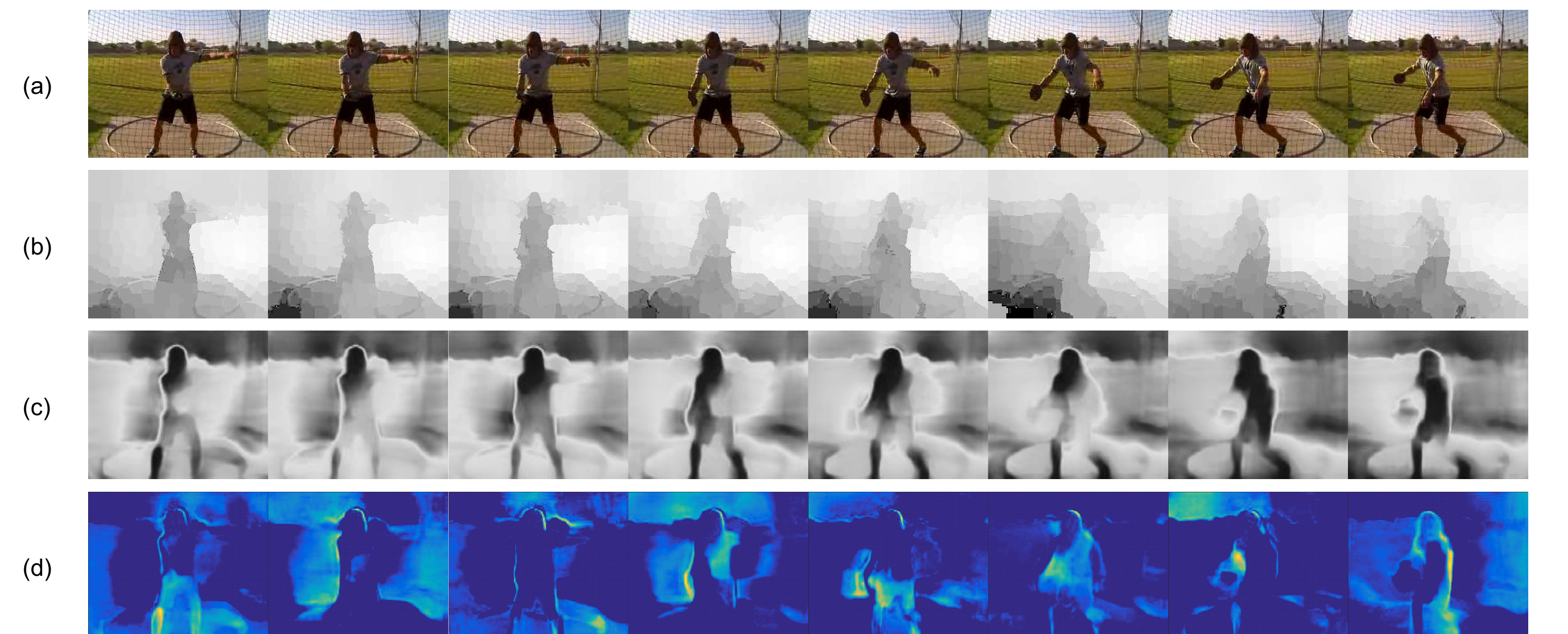}
	\caption{Depth maps estimated from the video v$\_$ThrowDiscus$\_$g05$\_$c02.avi in the UCF101 dataset. (a): raw RGB frames; (b): depth maps extracted using \cite{cnnfieldDepthLiu2015}; (c): depth maps extracted using \cite{eigenDepth2015}; (d): the absolute difference between consecutive depth maps in (c). Blue indicates smaller values and yellow larger ones. This figure is best viewed in color}
	\label{fig:depthMapExamples}
\end{figure}

\subsection{Depth Extraction}	
\label{subsec:depthExtraction}
Extracting depth maps from video has been studied for some time now \cite{zhangDepthVideo2009,multiviewDepth2010,depthGeometric2014}. Most approaches, however, are not applicable since they either require stereo video or additional information such as geometric priors. There are a few works \cite{miaomiao2015} which extract depth maps from monocular video alone but they are computationally too expensive which does not scale to problems like ours.

We therefore turn to frame-by-frame depth extraction and enforce temporal consistency through a normalization step. Depth from images has made much progress recently \cite{cnnfieldDepthLiu2015,coupleDepth2015,Intrinsicdepth2015,eigenDepth2015} and is significantly more efficient for extracting depth from video. We consider two state-of-the-art approaches to extract depth from images, \cite{cnnfieldDepthLiu2015} and \cite{eigenDepth2015}, based on their accuracy and efficiency.

\noindent \textbf{Deep Convolutional Neural Fields (DCNF) \cite{cnnfieldDepthLiu2015}:}  This work jointly explores the capacity of deep ConvNets and continuous CRFs to estimate depth from an image. Depth is predicted through maximum a posterior (MAP) inference which has a closed-form solution.
We apply the implementation kindly provided by the authors \cite{cnnfieldDepthLiu2015} but discard the time consuming ``inpainting'' procedure which is not important for our application. Our modified implementation takes only $0.09$s per frame to extract a depth map.

\noindent \textbf{Multi-scale Deep Network \cite{eigenDepth2015}:} Unlike DCNF above, this method does not utilize super-pixels and thus results in smoother depth maps. It uses a sequence of scales to progressively refine the predictions and to capture image details both globally and locally. Although the model can also be used to predict surface normals and semantic labels within a common ConvNets architecture, we only use it to extract depth maps. Our modified implementation takes only $0.01$s per frame to extract a depth map.

Fig. \ref{fig:depthMapExamples} visually compares the per-frame depths maps generated by the two approaches. 
We observe that 1) \cite{eigenDepth2015} (Fig. \ref{fig:depthMapExamples}c) results in smoother maps since it does not utilize super-pixels like \cite{cnnfieldDepthLiu2015} (Fig. \ref{fig:depthMapExamples}b), and 2) \cite{eigenDepth2015} preserves structural details, such as the border between the sky and the trees, better than \cite{cnnfieldDepthLiu2015} due to its multi-scale refinement. 
An ablation study (see supplemental materials) shows \cite{eigenDepth2015} results in better action recognition performance so we use it to extract per-frame depth maps for the rest of the paper.

\subsection{Spatio-Temporal Depth Normalization}
\label{subsec:preprocessing}
We now have depth sequences. While this makes our problem similar to work on action recognition from depth-sensor data such as \cite{pichaowang2015}, these methods are not applicable for a number of reasons. First, their inputs are point clouds which allows them to derive depth sequences from multiple perspectives for a single video as well as augment their training data through virtual camera movement. We only have a single fixed viewpoint. Second, their depth information has much higher fidelity since it was acquired with a depth sensor. Ours is prohibitively noisy to use a single 2D depth motion map to represent an entire video as is done in \cite{pichaowang2015}. We must develop new methods.

The first step is to reduce the noise by enforcing temporal consistency under the assumption that depth does not change significantly between frames. We introduce a temporal normalization scheme which constrains the furthest part of the scene to remain approximately the same throughout a clip. We find this works best when applied separately to three horizontal spatial windows and so we term the method spatio-temporal depth normalization (STDN). Specifically, let $\mathbf{x}$ be a frame. We then take $n$ consecutive frames $[\mathbf{x}_{t1}, \mathbf{x}_{t2}, \dots, \mathbf{x}_{tn}]$ to form a volume (clip) which is divided spatially into three equal-sized subvolumes that represent the top, middle, and bottom parts~\cite{Oneata2013}. We take the $95$\textsuperscript{th} percentile of the depth distribution as the furthest scene element in each subvolume. The $95$\textsuperscript{th} percentile of the corresponding window in each frame is then linearly scaled to equal this furthest distance.

We also investigated other methods to enforce temporal consistency including intra-frame normalization, temporal averaging (uniform as well as Gaussian) with varying temporal window sizes, and warping. None performed as well as the proposed STDN (see supplemental materials). 

\subsection{ConvNets Architecture Selection}
\label{subsec:convNets}
Recent progress in action recognition based on ConvNets can be attributed to two models: a two-stream approach based on 2D ConvNets \cite{twostream2014,wanggoodpractice2015} which separately models the spatial and temporal information, and 3D ConvNets which jointly learn spatio-temporal features \cite{3dconv2012,c3d2015}. These models are applied to RGB video sequences. We explore and adapt them for our depth sequences.

\noindent \textbf{2D ConvNets:}
In \cite{twostream2014}, the authors compute a spatial stream by adapting 2D ConvNets from image classification~\cite{alexnet2012} to action recognition. We do the same here except we use depth sequences instead of RGB video sequences. We term this our \emph{depth-spatial stream} to distinguish it from the standard spatial stream which we will refer to as RGB-spatial stream for clarity. Our depth-spatial stream is pre-trained on the ILSVRC-2012 dataset \cite{ILSVRC15} with the VGG-$16$ implementation \cite{vgg16192015} and fine-tuned on our depth sequences. \cite{twostream2014} also computes a temporal stream by applying 2D ConvNets to optical flow derived from the RGB video. We could similarly compute optical flow from our depth sequences but this would be redundant (and very noisy) so we instead propose a different depth-temporal stream below in section \ref{sec:DepthFlow}.

\noindent \textbf{3D ConvNets:} In \cite{3dconv2012,c3d2015}, the authors show that 2D ConvNets ``forget'' the temporal information in the input signal after every convolution operation. They propose 3D ConvNets which analyze sets of contiguous video frames organized as clips. We apply this approach to clips of depth sequences. We term this \emph{depth-C3D} to distinguish it from the standard 3D ConvNets which we will refer to as RGB-C3D for clarity. Our depth-C3D net is pre-trained using the Sports-1M dataset \cite{KarpathyCVPR14} and fine-tuned on our depth sequences.

\subsection{Depth-Temporal Stream}
\label{sec:DepthFlow}
Here, we look to augment our depth-spatial stream with a depth-temporal stream. We take inspiration from work on action recognition from depth-sensor data and adapt depth motion maps~\cite{DMMHOG2012} to our problem. In~\cite{DMMHOG2012}, a single 2D DMM is computed for an entire sequence by thresholding the difference between consecutive depth maps to get per-frame (binary) motion energy and then summing this energy over the entire video. A 2D DMM summarizes where depth motion occurs.

We instead calculate the motion energy as the absolute difference between consecutive depth maps \emph{without thresholding} in order to retain the subtle motion information embedded in our noisy depth sequences. We also accumulate the motion energy over clips instead of entire sequences since the videos in our dataset are longer and less-constrained compared to the depth-sensor sequences in \cite{3Dactiondepth2010,actionpairsHON4D2013,MSR3Dactivity,MSR3Dgesture,3DonlineAction} and so our depth sequences are too noisy to be summarized over long periods. In many cases, the background would simply dominate.

We compute one modified depth motion map (MDMM) for a clip of $N$ depth maps as
\begin{equation}
	\text{MDMM}_{t_{start}} = \sum_{t_{start}}^{t_{start}+N} | \text{map}^{t_{start}+1} - \text{map}^{t_{start}}|,
	\label{eq:MDMM}
\end{equation}
where $t_{start}$ is the first frame of the clip, $N$ is the duration of the clip, and $\text{map}^{t}$ is the depth map at frame $t$. Multiple MDMMs are computed for each video. Each MDMM is then input to a 2D ConvNet for classification. We term this our \emph{depth-temporal stream}. We combine it with our depth-spatial stream to create our \emph{depth two-stream} (see Fig. \ref{fig:workflow}). Similar to the depth-spatial stream, the depth-temporal stream is pre-trained on the ILSVRC-2012 dataset \cite{ILSVRC15} with the VGG-$16$ network \cite{vgg16192015} and fine-tuned on the MDMMs.

We also consider a simpler temporal stream by taking the absolute difference between adjacent depth maps and inputting this difference sequence to a 2D ConvNet. We term this our \emph{baseline depth-temporal stream}. Fig. \ref{fig:depthMapExamples}d shows an example sequence of this difference. It does a good job at highlighting changes in the depth despite the noisiness of the image-based depth estimation.

\subsection{ConvNets: Feature Extraction or End-to-End Classification}
\label{sec:earlyLateFusion}
The ConvNets in our depth two-stream and depth-C3D models default to end-to-end classifiers. We investigate whether to use them instead as feature extractors followed by SVM classifiers. This also allows us to investigate early versus late fusion. We use our depth-spatial stream for illustration.

Features are extracted from two layers of our fine-tuned ConvNets. We extract the activations of the first fully-connected layer (\textsf{fc6}) on a per-frame basis. These are then averaged over the entire video and L$2$-normalized to form a $4096$-dim video-level descriptor. We also extract activations from the convolutional layers as they contain spatial information. We choose the \textsf{conv5} layer, whose feature dimension is $7 \times 7 \times 512$ ($7$ is the size of the filtered images of the convolutional layer and $512$ is the number of convolutional filters).
By considering each convolutional filter as a latent concept, the \textsf{conv5} features can be converted into $7^{2}$ latent concept descriptors (LCD) \cite{LCDXu2015} of dimension $512$. We also adopt a spatial pyramid pooling (SPP) strategy \cite{sppnet} similar to \cite{LCDXu2015}. We apply principle component analysis (PCA) to de-correlate and reduce the dimension of the LCD features to $64$ and then encode them using vectors of locally aggregated descriptors (VLAD) \cite{VLAD}. This is followed by intra- and L2-normalization to form a $16384$-dim video-level descriptor. 

Early fusion consists of concatenating the \textsf{fc6} and \textsf{conv5} features for input to a single multi-class linear SVM classifier \cite{liblinear} (see Fig. \ref{fig:workflow}). Late fusion consists of feeding the features to two separate SVM classifiers and computing a weighted average of their probabilities. The optimal weights are selected by grid-search.

\section{Experiments}
\label{sec:experiments}
The goal of our experiments is two-fold. First, to explore the various design options described in section \ref{sec:methodology} Methodology. Second, to show that our depth2action framework is complementary to standard approaches to large-scale action recognition based on appearance and translational motion and achieves state-of-the-art results when combined with them.

\subsection{Datasets}
\label{sec:dataset}
We perform experiments on three widely-used publicly-available action recognition benchmark datasets, UCF101 \cite{ucf101}, HMDB51 \cite{hmdb51}, and ActivityNet \cite{activityNet}.

\textbf{UCF101} is composed of realistic action videos from YouTube. It contains $13320$ videos in $101$ action classes. It is one of the most popular benchmark datasets because of its diversity in terms of actions and the presence of large variations in camera motion, object appearance and pose, object scale, viewpoint, cluttered background, illumination conditions, etc. 
%All the videos have a fixed frame rate of $25$ fps with a spatial resolution of $320 \times 240$ pixels.
\textbf{HMDB51} is composed of $6766$ videos in $51$ action classes extracted from a wide range of sources.
It contains both original videos as well as stabilized ones, but we only use the original videos. 
Both UCF101 and HMDB51 have a standard three split evaluation protocol and we report the average recognition accuracy over the three training and test splits.
As suggested by the authors in \cite{activityNet}, we use \textbf{ActivityNet} release 1.2 for our experiments due to the noisy crowdsourced labels in release 1.1. 
The second release consists of $4819$ training, $2383$ validation, and $2480$ test videos in $100$ activity classes. 
Though the number of videos and classes are similar to UCF101, ActivityNet is a much more challenging benchmark because it has greater intra-class variance and consists of longer, untrimmed videos.
The evaluation metric we used in this paper is top-1 accuracy for all three datasets.

\begin{table}[t]
\begin{center}
\caption{Recognition performance of our proposed configurations on three benchmark datasets. (a): Our spatio-temporal depth normalization (STDN) indicated by (N) is shown to improve performance for all configurations on all datasets. (b): Using the ConvNets to extract features is better than using them as end-to-end classifiers. Also, early fusion of features is better than late fusion of SVM probabilities. See the text for discussion on depth two-stream versus depth-C3D}
\label{tab:result1}
\begin{minipage}{0.5\textwidth}%
	\centering
	\subfloat[][Effectiveness of STDN]{
		\scalebox{0.68}{
			\begin{tabular}{| c | c | c | c |}
				\hline
				Model								& 	 UCF101				&  HMDB51 				&  ActivityNet					\\
				\hline		
				Depth-Spatial						& 	$58.8\%$ 			& $37.9\%$				&	$35.9\%$					\\	
				Depth-Spatial (N)					& 	$59.1\%$ 			& $38.3\%$				&	$36.4\%$					\\
				Depth-Temporal Baseline				& 	$61.8\%$ 			& $40.6\%$				&	$38.2\%$					\\
				Depth-Temporal Baseline (N) 		& 	$63.3\%$ 			& $42.0\%$				&	$39.8\%$					\\			
				Depth-Temporal 						& 	$63.9\%$ 			& $42.6\%$				&	$39.7\%$					\\	
				Depth-Temporal (N) 					& 	$65.1\%$ 			& $43.5\%$				&	$40.9\%$					\\	
				Depth Two-Stream  					&   $65.6\%$ 			& $44.2\%$				&	$42.7\%$					\\
				Depth Two-Stream (N)  				&   $\textbf{67.0\%}$ 	& $\textbf{45.4\%}$		&	$44.2\%$					\\
				\hline
				\hline
				Depth-C3D 							&   $61.7\%$ 			& $40.9\%$				&	$45.9\%$					\\
				Depth-C3D (N)						&   $63.8\%$ 			& $42.8\%$				&	$\textbf{47.4\%}$			\\
				\hline
			\end{tabular}
		}
	}
\end{minipage}%
%\qquad
\begin{minipage}{0.5\textwidth}%
	\centering
	\subfloat[][Features or End-to-End Classifier ]{
		\scalebox{0.68}{
			\begin{tabular}{| c | c | c | c |}
				\hline
				Model								& 	UCF101				& HMDB51 				& ActivityNet					\\
				\hline			
				Depth Two-Stream 					& 	$67.0\%$ 			& $45.4\%$				&	$44.2\%$					\\	
				Depth Two-Stream \textsf{fc6}		& 	$68.2\%$ 			& $46.5\%$				&	$45.3\%$					\\	
				Depth Two-Stream \textsf{conv5}		& 	$70.1\%$ 			& $48.2\%$				&	$47.0\%$					\\	
				Depth Two-Stream Early				& 	$\textbf{72.5\%}$ 	& $\textbf{49.7\%}$ 	&	$49.6\%$					\\	
				Depth Two-Stream Late				& 	$70.9\%$ 			& $48.9\%$				&	$48.7\%$					\\			
				\hline
				\hline
				Depth-C3D  							&   $63.8\%$ 			& $42.8\%$				&	$47.4\%$					\\
				Depth-C3D \textsf{fc6}				&   $64.9\%$ 			& $43.9\%$				&	$47.9\%$					\\
				Depth-C3D \textsf{conv5b}			&   $66.7\%$ 			& $45.0\%$				&	$49.1\%$					\\
				Depth-C3D Early						&   $69.5\%$ 			& $46.6\%$				&	$\textbf{52.1\%}$			\\
				Depth-C3D Late						&   $67.8\%$ 			& $45.7\%$				&	$51.0\%$					\\
				\hline
			\end{tabular}
		}
	}
\end{minipage}
\end{center}
\end{table} 

\subsection{Implementation Details}
\label{sec:implementationDetails}
We use the Caffe toolbox \cite{jia2014caffe} to implement the ConvNets.
The network weights are learned using mini-batch stochastic gradient descent ($256$ frames for two-stream ConvNets and $30$ clips for 3D ConvNets) with momentum (set to $0.9$).

\noindent \textbf{Depth Two-Stream:} 
We adapt the VGG-$16$ architecture \cite{vgg16192015} and use ImageNet models as the initialization for both the depth-spatial and depth-temporal net training.
As in \cite{wanggoodpractice2015}, we adopt data augmentation techniques such as corner cropping, multi-scale cropping, horizontal flipping, etc. to help prevent overfitting, as well as high dropout ratios ($0.9$ and $0.8$ for the fully connected layers).  
The input to the depth-spatial net is the per-frame depth maps, while the input to the depth-temporal net is either the depth difference between adjacent frames (in the baseline case) or the MDMMs.
For generating the MDMMs, we set $N$ in equation \ref{eq:MDMM} to $10$ frames as a subvolume.
For the depth-spatial net, the learning rate decreases from $0.001$ to $1/10$ of its value every $15$K iterations, and the training stops after $66$K iterations. 
For the depth-temporal net, the learning rate starts at $0.005$, decreases to $1/10$ of its value every $20$K iterations, and the training stops after $100$K iterations. 

\noindent \textbf{Depth-C3D:} We adopt the same architecture as in \cite{c3d2015}. 
The Depth-C3D net is pre-trained on the Sports-1M dataset \cite{KarpathyCVPR14} and fine-tuned on estimated depth sequences. 
During fine-tuning, the learning rate is initialized to $0.005$, decreased to $1/10$ of its value every $8$K iterations, and the training stops after $34$K iterations. 
Dropout is applied with a ratio of $0.5$. 

Note that since the number of training videos in the HMDB51 dataset is relatively small, we use ConvNets fine-tuned on UCF101, except for the last layer, as the initialization (for both 2D and 3D ConvNets). The fine-tuning stage starts with a learning rate of $10^{-5}$ and converges in one epoch.

\subsection{Results}
\label{sec:results}
\textbf{Effectiveness of STDN:}  
Table \ref{tab:result1}(a) shows the performance gains due to our proposed normalization. STDN improves recognition performance for all approaches on all datasets. The gain is typically around 1-2\%. We set the normalization window (n in section \ref{subsec:preprocessing}) to $16$ frames for UCF101 and ActivityNet, and $8$ frames for HMDB51. We further observe that (i) Depth-C3D benefits from STDN more than depth two-stream. This is possibly because the input to depth-C3D is a 3D volume of depth sequences while the input to depth two-stream is the individual depth maps. Temporal consistency is important for the 3D volume.
(ii) Depth-temporal benefits from STDN more than depth-spatial. This is expected since the goal of the normalization is to improve the temporal consistency of the depth sequences and only the depth-temporal stream ``sees'' multiple depth-maps at a time. From now on, all results are based on depth sequences that have been normalized.

\noindent \textbf{Depth Two-Stream versus Depth-C3D:}
As shown in Table \ref{tab:result1}(a), depth two-stream performs better than depth-C3D for UCF101 and HMDB51, while the opposite is true for ActivityNet. 
This suggests that depth-C3D may be more suitable for large-scale video analysis. Though the second release of ActivityNet has a similar number of action clips as UCF101, in general, the video duration is much ($30$ times) longer than that of UCF101. Similar results for 3D ConvNets versus 2D ConvNets was observed in \cite{DAP3D2016}. The computational efficiency of depth-C3D also makes it more suitable for large-scale analysis. Although our depth-temporal net is much faster than the RGB-temporal net (which requires costly optical flow computation), depth-two stream is still significantly slower than depth-C3D. We therefore recommend using depth-C3D for large-scale applications.

\begin{figure}[t]
	\centering
	\subfloat[]{\includegraphics[width=0.495\linewidth, trim=0 0 0 20,clip]{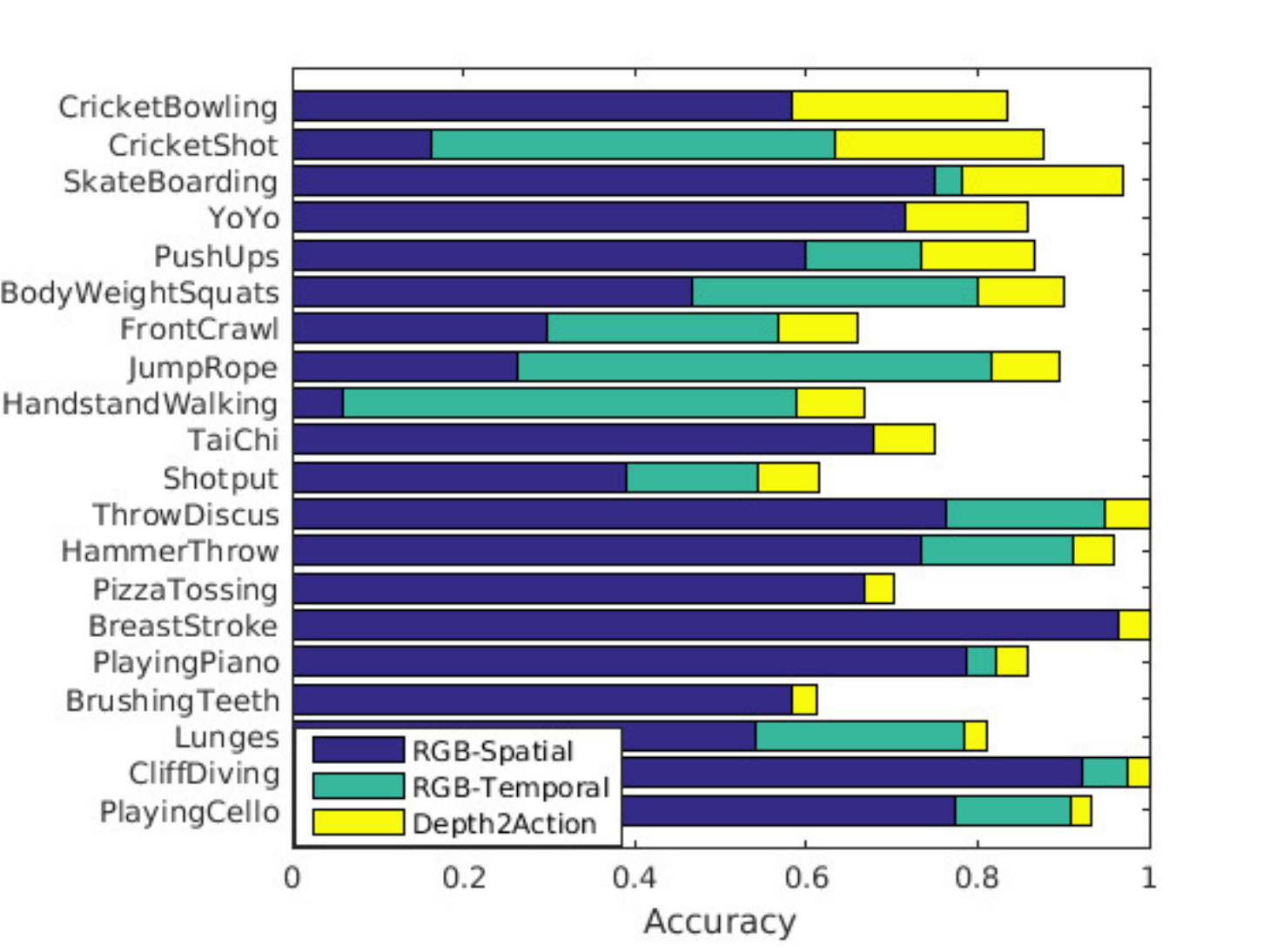}\label{fig:depthimprove}}\hspace{0pt}
	\subfloat[]{\includegraphics[width=0.495\linewidth, trim=0 0 0 0,clip]{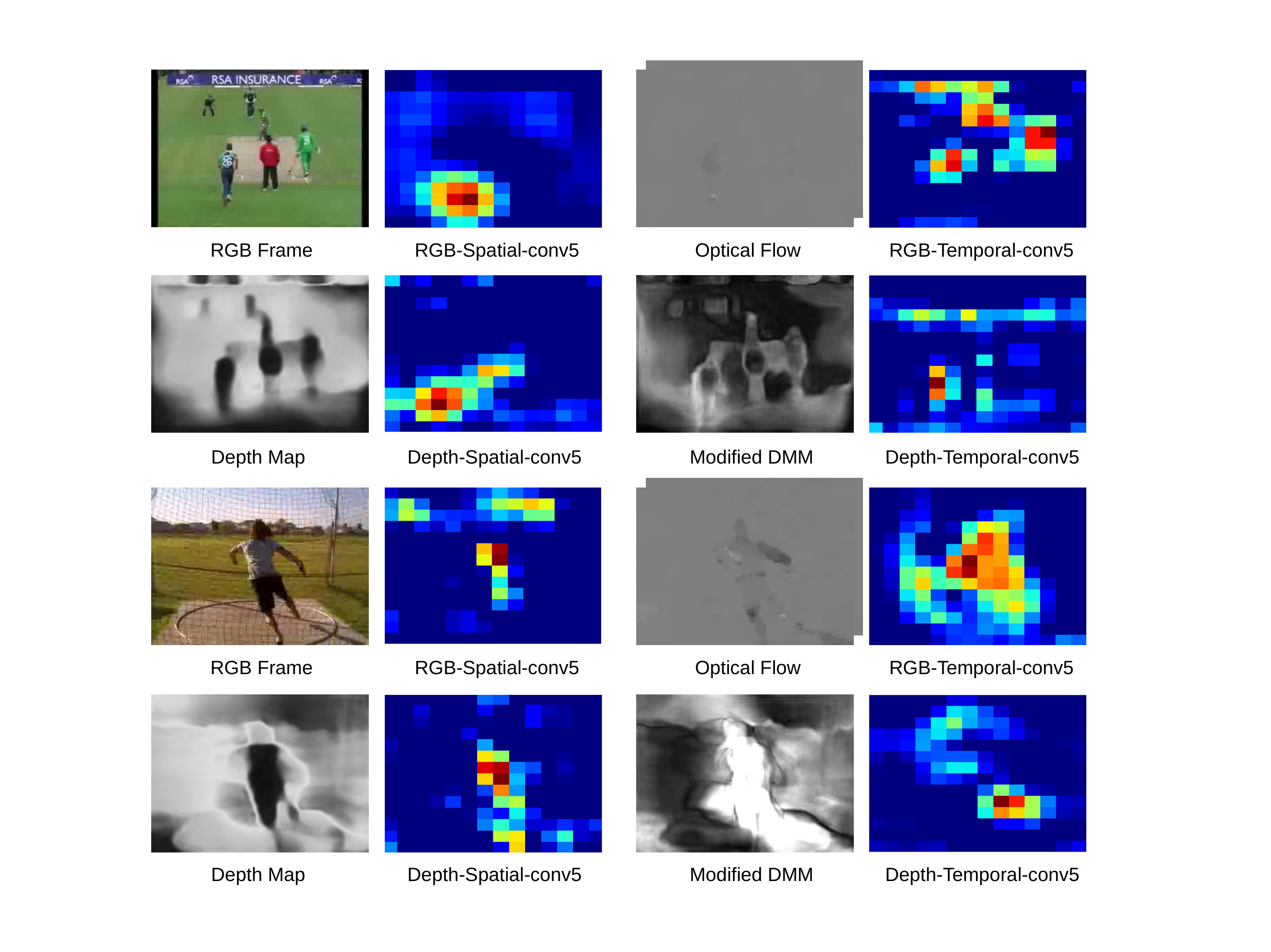}\label{fig:conv_filters}}
	\caption{(a) Recognition results on the first split of UCF101. Plot showing the classes for which our proposed depth2action framework (yellow) outperforms RGB-spatial (blue) and RGB-temporal (green) streams. (b) Visualizing the convolutional feature maps of four models: RGB-spatial, RGB-temporal, depth-spatial, and depth-temporal. Pairs of inputs and resulting feature maps are shown for each model for two actions, ``CriketBowling'' and ``ThrowDiscus''. This figure is best viewed in color}
	\label{fig:depthBetter}
\end{figure}

\noindent \textbf{ConvNets for Feature Extraction versus End-to-End Classification:}
Table \ref{tab:result1}(b) shows that treating the ConvNets as feature extractors performs significantly better than using them for end-to-end classification. This agrees with the observations of others~\cite{BallasDeeper2015,c3d2015,shichaoPooling2015}. We further observe that the VLAD encoded \textsf{conv5} features perform better than \textsf{fc6}. This improvement is likely due to the additional discriminative power provided by the spatial information embedded in the convolutional layers. Another attractive property of using feature representations is that we can manipulate them in various ways to further improve the performance. 
For instance, we can employ different (i) encoding methods: Fisher vector~\cite{Oneata2013}, VideoDarwin \cite{videoDarwin}; (ii) normalization techniques: rank normalization \cite{rankingReranking2015}; and (iii) pooling methods: line pooling \cite{shichaoPooling2015}, trajectory pooling \cite{tddwang2015,shichaoPooling2015}, etc. 

\noindent \textbf{Early versus Late Fusion:}
Table \ref{tab:result1}(b) also shows that early fusion of features through concatenation performs better than late fusion of SVM probabilities. Late fusion not only results in a performance drop of around $1.0\%$ but also requires a more complex processing pipeline since multiple SVM classifiers need to be trained. UCF101 benefits from early fusion more than the other two datasets. This might be due to the fact that UCF101 is a trimmed video dataset and so the content of individual videos varies less than in the other two datasets. Early fusion of multiple layers' activations is typically more robust to noisy data.

\noindent \textbf{Depth2Action:}
We thus settle on our proposed depth2action framework. For medium-scale video datasets like UCF101 and HMDB51, we perform early fusion of \textsf{conv5} and \textsf{fc6} features extracted using a depth two-stream configuration. For large-scale video datasets like ActivityNet, we perform early fusion of \textsf{conv5b} and \textsf{fc6} features extracted using a depth-C3D configuration. These two models are shown in Fig. \ref{fig:workflow}.

\begin{figure}[t]
	\centering
	\includegraphics[width=1.0\linewidth,trim=0 360 0 0,clip]{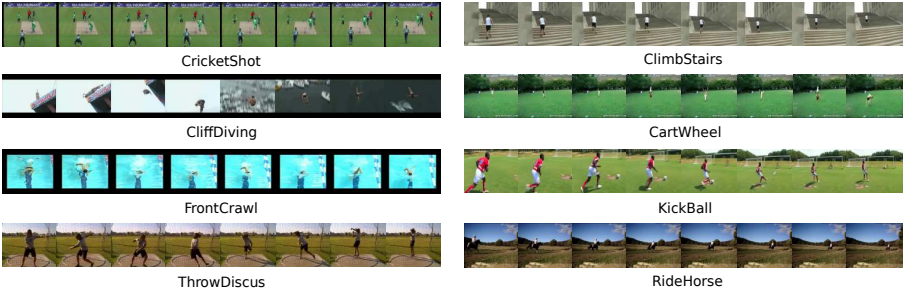}
	\caption{Sample video frames of action classes that benefit from depth information. Left: UCF101. Right: HMDB51. This figure is best viewed in color}
	\label{fig:demonstration}
\end{figure}

\subsection{Discussion}
\label{sec:discussion}
\noindent \textbf{Class-Specific Results:}
We investigate the specific classes for which depth information is important. To do this, we compare the per-class performance of our depth2action framework with standard methods that use appearance and translational motion in the RGB domain. We first compute the performances of an RGB-spatial stream which takes the RGB video frames as input and an RGB-temporal stream which takes optical flow (computed in the RGB domain) as input. We then identify the classes for which our depth2action performs better than both the RGB-spatial and RGB-temporal streams. We compute these results for the first split of the UCF101 dataset. Fig. \ref{fig:depthimprove} shows the $20$ classes for which our depth2action framework performs best (in order of decreasing improvement). For example, for the class CricketShot, RGB-spatial achieves an accuracy of around $0.18$, RGB-temporal achieves around $0.62$, while our depth2action achieves around 0.88. (For those classes where RGB-spatial performs better than RGB-temporal, we simply do not show the performance of RGB-temporal.) Depth2action clearly represents a complementary approach especially for classes where the RGB-spatial and RGB-temporal streams perform relatively poorly such as CriketBowling, CriketShot, FrontCrawl, HammerThrow, and HandStandWalking. Recall from Fig. \ref{fig:depthIsUseful} that CriketBowling and CriketShot are very similar with respect to appearance and translational motion. These are shown the be the two classes for which depth2action provides the most improvement, achieving respectable accuracies of above $0.8$.

Sample video frames from classes in the UCF101 (left) and HMDB51 (right) datasets which benefit from depth information are show in Fig. \ref{fig:demonstration} (see supplemental materials for more samples).

\noindent \textbf{Visualizing Depth2Action:}
We visualize the convolutional feature maps (\textsf{conv5}) to better understand how depth2action encodes depth information and how this encoding is different from that of RGB two-stream models. Fig. \ref{fig:conv_filters} shows pairs of inputs and resulting feature maps for four models: RGB-spatial, RGB-temporal, depth-spatial, and depth-temporal. (The feature maps are displayed using a standard heat map in which warmer colors indicate larger values.) The top four pairs are for ``CriketBowling'' and bottom four pairs are for ``ThrowDiscus'' (see supplemental materials for more action classes).

In general, the depth feature maps are sparser and more accurate than the RGB feature maps, especially for the temporal streams. The depth-spatial stream correctly encodes the bowler and the batter in ``CriketBowling'' and the discus thrower in ``ThrowDiscus'' as being salient while the RGB-stream gets distracted by other parts of the scene. The depth-temporal stream clearly identifies the progression of the bowler into the scene in ``CriketBowling'' and the movement of the discus thrower's leg into the scene in ``ThrowDiscus'' as being salient while the RGB-temporal stream is distracted by translational movement throughout the scene. These results demonstrate that our proposed depth2action approach does indeed focus on the correct regions in classes for which depth is important.

\begin{table}[t]
\centering
\caption{Comparison of RGB two-stream, IDT computed from RGB video, and depth2action, and their combinations for the UCF101 dataset. $\Delta$ indicates the performance increase with respect to RGB two-stream taken as the baseline}
\label{tab:result2}
\scalebox{0.80}{
	\begin{tabular}{ c | c | c | c | c| c}
		\hline
		Model							& 	split01		&  split02 	&  split03  &  Average		&$\Delta$ 		\\
		\hline		
		RGB Two-Stream Baseline 			&   $90.8\%$ 	& $92.0\%$	&  $91.3\%$	&  $91.4\%$		&  $0$			\\
		IDT								&   $83.1\%$ 	& $85.9\%$	&  $85.1\%$	&  $84.7\%$		&  -			\\
		Depth2Action					&   $71.9\%$ 	& $73.0\%$	&  $72.5\%$	&  $72.5\%$		&  -			\\
		\hline
		RGB Two-Stream+IDT 					&   $91.9\%$ 	& $92.6\%$	&  $91.8\%$	&  $92.1\%$		&  $0.8\%$		\\
		RGB Two-Stream+Depth2Action 		&   $91.7\%$ 	& $92.5\%$	&  $91.8\%$	&  $92.0\%$		&  $0.7\%$		\\
		Depth2Action+IDT 				&   $84.3\%$ 	& $86.6\%$	&  $86.4\%$	&  $85.8\%$		&  -			\\
		\hline
		RGB Two-Stream+Depth2Action+IDT		&   $92.5\%$ 	& $93.8\%$	&  $92.8\%$	&  $93.0\%$		&  $1.8\%$		\\	
		\hline
	\end{tabular}
}
\end{table} 

\noindent \textbf{What about IDT?}
We compare our depth2action framework with improved dense trajectories (IDT) computed from RGB video. IDT has been shown to be the best hand-crafted features for action recognition~\cite{idtfWang2013}. It is known to perform well under various camera motions (e.g. pan, tilt and zoom) and zoom can be considered global depth change.

While the top part of Table \ref{tab:result2} shows that IDT outperforms depth2action, which is not surprising due to how noisy our estimated depth maps are, we turn our attention to the performance obtained by combining these two approaches with an RGB two-stream model. Rows four and five show that the performance achieved by combining depth2action with RGB two-stream is on par with the combination of IDT with RGB two-stream (and both perform significantly better than IDT). The last column shows the improvement over RGB two-stream alone. This demonstrates that although depth2action is not as effective as IDT when taken alone, \emph{it is as complementary to RGB two-stream as IDT}. This point is even more significant given the fact that IDT requires several orders of magnitude more computation time and storage space (mainly to extract and store the features) than depth2action. The combination of depth2action and RGB two-stream is much preferred over that of IDT and RGB two-stream for large-scale analysis.
%Our method only need to calculate the depth images and fine-tune one deep ConvNet.
The last row of Table \ref{tab:result2} shows the results of combining all three approaches. We again get an improvement. This result turns out to be state-of-the-art for this dataset and stresses the importance and complementarity of jointly exploiting appearance, translational motion, and depth for action recognition.

\subsection{Comparison with State-of-the-art}
\label{sec:stateoftheart}
\begin{table}[t]
\caption{Comparison with the state-of-the-art. $^{\ast}$ indicates the results are from our implementation of the method. Two-stream and C3D here  is RGB based \label{tab:result3}}
\parbox{1.0\linewidth}{
\centering
\scalebox{0.7}{
\begin{tabular}{ c | c || c | c || c | c }
	\hline
	Algorithm	& 	UCF101		&  	Algorithm 	&  HMDB51   &  	Algorithm 		&  ActivityNet\\
	\hline		
	Srivastava et al. \cite{unsupervisedLSTM2015} &   $84.3\%$ 	& Srivastava et al. \cite{unsupervisedLSTM2015}	&  $44.1\%$	& Wang $\text{\&}$ Schmid \cite{idtfWang2013} &   $61.3\%^{\ast}$  \\
	
	Wang $\text{\&}$ Schmid \cite{idtfWang2013}	  &  $85.9\%$	& 	Oneata et al. \cite{Oneata2013}				&  $54.8\%$	& Simonyan $\text{\&}$ Zisserman \cite{twostream2014}  &   $67.1\%^{\ast}$	\\	
	
	Simonyan $\text{\&}$ Zisserman \cite{twostream2014} 	&   $88.0\%$ 	& Wang $\text{\&}$ Schmid \cite{idtfWang2013}	&  $57.2\%$	& Tran et al. \cite{c3d2015}  &		$69.4\%^{\ast}$\\
	
	Jain et al. \cite{15000object2015} 	&   $88.5\%$ 	& 	Simonyan $\text{\&}$ Zisserman \cite{twostream2014}			&  $59.1\%$	& &\\
	
	Ng et al. \cite{beyondshort2015}   		&   $88.6\%$ 	& 	Sun et al. \cite{factorized3DCNNSun2015}		&  $59.1\%$	& &\\
	Lan et al. \cite{MIFS2015}				&   $89.1\%$ 	& 	Jain et al. \cite{15000object2015}				&  $61.4\%$	& &\\
	Zha et al. \cite{imageCNNvideo2015}   	&   $89.6\%$ 	& 	Fernando et al. \cite{videoDarwin}		&  $63.7\%$	& &\\
	Tran et al. \cite{c3d2015} 				&   $90.4\%$ 	& 	Lan et al. \cite{MIFS2015}					&  $65.1\%$	& &\\
	Wu et al. \cite{hybridWu2015}			&   $91.3\%$ 	& 	Wang et al. \cite{tddwang2015}				&  $65.9\%$	& &\\
	Wang et al. \cite{tddwang2015}			&   $91.5\%$ 	& 	Peng et al. \cite{StackedFV2014}	 & 	$66.8\%$ & &\\		
	\hline
	Depth2Action 							&   $72.5\%$	& 	Depth2Action	&  $49.7\%$	 & Depth2Action  	&	$52.1\%$	 \\
	+Two-Stream				&   $92.0\%$	& 	+Two-Stream	&  $67.1\%$	& +C3D &  $71.2\%$  \\
	+IDT+Two-Stream 				&   $\mathbf{93.0\%}$	& 	+IDT+Two-Stream	&  $\mathbf{68.2\%}$	& +IDT+C3D &  $\mathbf{73.4\%}$  \\
	\hline
\end{tabular}
}
}
\end{table} 
Table \ref{tab:result3} compares our approach with a large number of recent state-of-the-art published results on the three benchmarks. For UCF101 and HMDB51, the reported performance is the mean recognition accuracy over the standard three splits. The last row shows the performance of combining depth2action with RGB two-stream for UCF101 and HMDB51, and RGB C3D for ActivityNet, and also IDT features. We achieve state-of-the-art results on all three datasets through this combination, again stressing the importance of appearance, motion, and depth for action recognition.

We note that since there are no published results\footnote{The up-to-date leaderboard is at http://activity-net.org/evaluation.html.} for release 1.2 of ActivityNet, we report the results from our implementations of IDT \cite{idtfWang2013}, RGB two-stream \cite{twostream2014} and RGB C3D \cite{c3d2015}. 

\section{Conclusion}
\label{sec:conclusion}
We introduced \emph{depth2action}, the first investigation into depth for large-scale human action recognition where the depth cues are derived from the videos themselves rather than obtained using a depth sensor. This greatly expands the applicability of the method. Depth is estimated on a per-frame basis for efficiency and temporal consistency is enforced through a novel normalization step. Temporal depth information is captured using modified depth motion maps. A wide variety of design options are explored. Depth2action is shown to be complementary to standard approaches based on appearance and translational motion, and achieves state-of-the-art performance on three benchmark datasets when combined with them.

In addition to advancing state-of-the-art performance, the depth2action framework is a rich research problem. It bridges the gap between the RGB- and RGB-D-based action recognition communities. It consists of numerous interesting sub-problems such as fine-grained action categorization, depth estimation from single images/video, learning from noisy data, etc. The estimated depth information could also be used for other applications such as object detection/segmentation, event recognition, and scene classification. We will make our trained models and estimated depth maps publicly available for future research.

\noindent \textbf{Acknowledgements.} This work was funded in part by a National Science Foundation CAREER grant, $\#$IIS-1150115, and a seed grant from the Center for Information Technology in the Interest of Society (CITRIS). We gratefully acknowledge the support of NVIDIA Corporation through the donation of the Titan X GPU used in this work.

\bibliographystyle{splncs03}
\bibliography{egbib_eccv}
\end{document}